%% file: main.tex
\documentclass[a4paper,fleqn]{cas-dc}

\usepackage[numbers]{natbib}

\usepackage{url}
\usepackage[english]{babel} 
\usepackage{comment}
\usepackage{caption}
\usepackage{subcaption}
\usepackage{booktabs}
\usepackage{amsthm, amssymb}
\usepackage{amsmath}
\usepackage{textgreek}
\usepackage[scientific-notation=true]{siunitx}
\usepackage{threeparttablex}

\usepackage{graphicx}
\usepackage{algorithm}
\usepackage{algpseudocode}
\usepackage{multirow}
\usepackage{amsmath}
\usepackage{xcolor}
\usepackage{amssymb}
\usepackage{pifont}

\algnewcommand\Input{\item[\textbf{Input:}]}%
\algnewcommand\Output{\item[\textbf{Output:}]}%

\newcommand{\cmark}{\ding{51}}%
\newcommand{\xmark}{\ding{55}}%

\def\tsc#1{\csdef{#1}{\textsc{\lowercase{#1}}\xspace}}
\tsc{WGM}
\tsc{QE}
\tsc{EP}
\tsc{PMS}
\tsc{BEC}
\tsc{DE}

\begin{document}


\let\WriteBookmarks\relax
\def\floatpagepagefraction{1}
\def\textpagefraction{.001}

\shorttitle{PreMix: Label-Efficient MIL via Non-Contrastive Pre-training and Feature Mixing}

\shortauthors{Wong and Yi}

\title [mode = title]{PreMix: Label-Efficient Multiple Instance Learning via Non-Contrastive Pre-training and Feature Mixing}

\author[1]{Bryan Wong}[orcid=0000-0002-2257-3454]

\ead{bryan.wong@kaist.ac.kr}


\credit{Conceptualization, Methodology, Software, Validation, Formal analysis, Investigation, Writing - original draft, Writing - review \& editing, Visualization}

\affiliation[1]{organization={Graduate School of Data Science, KAIST},
    city={Daejeon},
    postcode={34141},
    country={Republic of Korea}}

\author[1, 2]{Mun Yong Yi}[orcid=0000-0003-1784-8983]

\cormark[1]

\ead{munyi@kaist.ac.kr}


\credit{Resources, Visualization, Writing - review \& editing, Supervision, Funding acquisition}

\affiliation[2]{organization={Department of Industrial and Systems Engineering, KAIST},
    city={Daejeon},
    postcode={34141},
    country={Republic of Korea}}

\cortext[cor1]{Corresponding author.}

\begin{abstract}
Multiple instance learning (MIL) has emerged as a powerful framework for weakly supervised whole slide image (WSI) classification, enabling slide-level predictions without requiring detailed patch-level annotations. Despite its success, a critical limitation of current MIL methods lies in the underutilization of pre-training for the MIL aggregator. Most existing approaches initialize the aggregator randomly and train it from scratch, making performance highly sensitive to the quantity of labeled WSIs and ignoring the abundance of unlabeled WSIs commonly available in clinical settings. To address this, we propose \textbf{PreMix}, a novel framework that leverages a non-contrastive pre-training method, Barlow Twins, augmented with the \textit{Slide Mixing} approach to generate additional positive pairs and enhance feature learning, particularly under limited labeled WSI conditions. Fine-tuning with Mixup and Manifold Mixup further enhances robustness by effectively handling the diverse sizes of gigapixel WSIs. Experimental results demonstrate that integrating PreMix as a plug-in module into HIPT yields an average F1 improvement of 4.7\% over the baseline HIPT across various WSI training sizes and datasets. These findings underscore its potential to advance WSI classification with limited labeled data and its applicability to real-world histopathology practices. The code is available at \url{https://github.com/bryanwong17/PreMix}
\end{abstract}






\maketitle

\section{Introduction}
\label{sec:Introduction}
\input{sections/1_Introduction}

\section{Related Work}
\label{sec:Related work}
\input{sections/2_Related_Work}

\section{Methods}
\label{sec:Methods}
\input{sections/3_Methods}

\section{Experiments}
\label{sec:Experiments}
\input{sections/4_Experiments}

\section{Discussion}
\label{sec:Discussion}
\input{sections/5_Discussion}

\section{Conclusions}
\label{sec:Conclusions}
\input{sections/6_Conclusions}

\printcredits

\section*{Declaration of competing interest}
The authors declare that they have no known competing financial interests or personal relationships that could have appeared to influence the work reported in this paper.

\section*{Acknowledgements}
This work was supported by the National Research Foundation of Korea (NRF) grant funded by the Korea government (MSIT) (No.RS-2022-NR068758).

\section*{Code availability}
The code is publicly available at \url{https://github.com/bryanwong17/PreMix}

\section*{Data availability}
The WSI datasets utilized in this study are publicly accessible through the following repositories: \href{https://ftp.cngb.org/pub/gigadb/pub/10.5524/100001_101000/100439/CAMELYON16/}{Camelyon16 dataset} and \href{https://wiki.cancerimagingarchive.net/display/Public/CPTAC+Pathology+Slide+Downloads}{CPTAC UCEC dataset}.

\section*{Declaration of generative AI and AI-assisted technologies in the writing process}

During the preparation of this work, the authors used an LLM (GPT-4o) to improve readability and language of the work. After using this tool, the authors reviewed and edited the content as needed and take full responsibility for the final version of the publication.

\bibliographystyle{model1-num-names}
\bibliography{bibliography}

\end{document}

%% file: sections/1_Introduction.tex
Histopathology provides microscopic insights into tissue structures, making it essential for diagnosing diseases, predicting outcomes, and guiding treatment \citep{Medical_Imaging_Opportunities}. Traditionally, this work has relied on the expertise of pathologists who examine tissue samples under a microscope, a process that can be time-consuming and subject to interpretation. In recent years, digital pathology has emerged as a major advancement, with whole slide images (WSIs) offering high-resolution digital scans of tissue slides \citep{WSI_Intro1}. These images provide a comprehensive view of tissue architecture and cellular details, helping clinicians make more accurate and consistent decisions \citep{WSI_Intro2, WSI_Intro3}.

Despite their potential, WSIs introduce new challenges. Their extremely large size, often reaching gigapixel resolution, makes it difficult to train deep learning models directly without dividing them into smaller patches due to hardware limitations. Moreover, annotating these images requires significant time and effort from medical professionals, and manual labeling can lead to inconsistencies across cases \citep{Annotation_Variability}. To address these issues, multiple instance learning (MIL) has gained attention as a weakly supervised approach for WSI classification. MIL enables models to learn from slide-level labels without needing detailed annotations for each patch, making it well-suited for clinical workflows where labeling resources are limited \citep{MIL_Framework}.

\begin{figure*}[!htb]
    \centering
    \includegraphics[width=\linewidth]{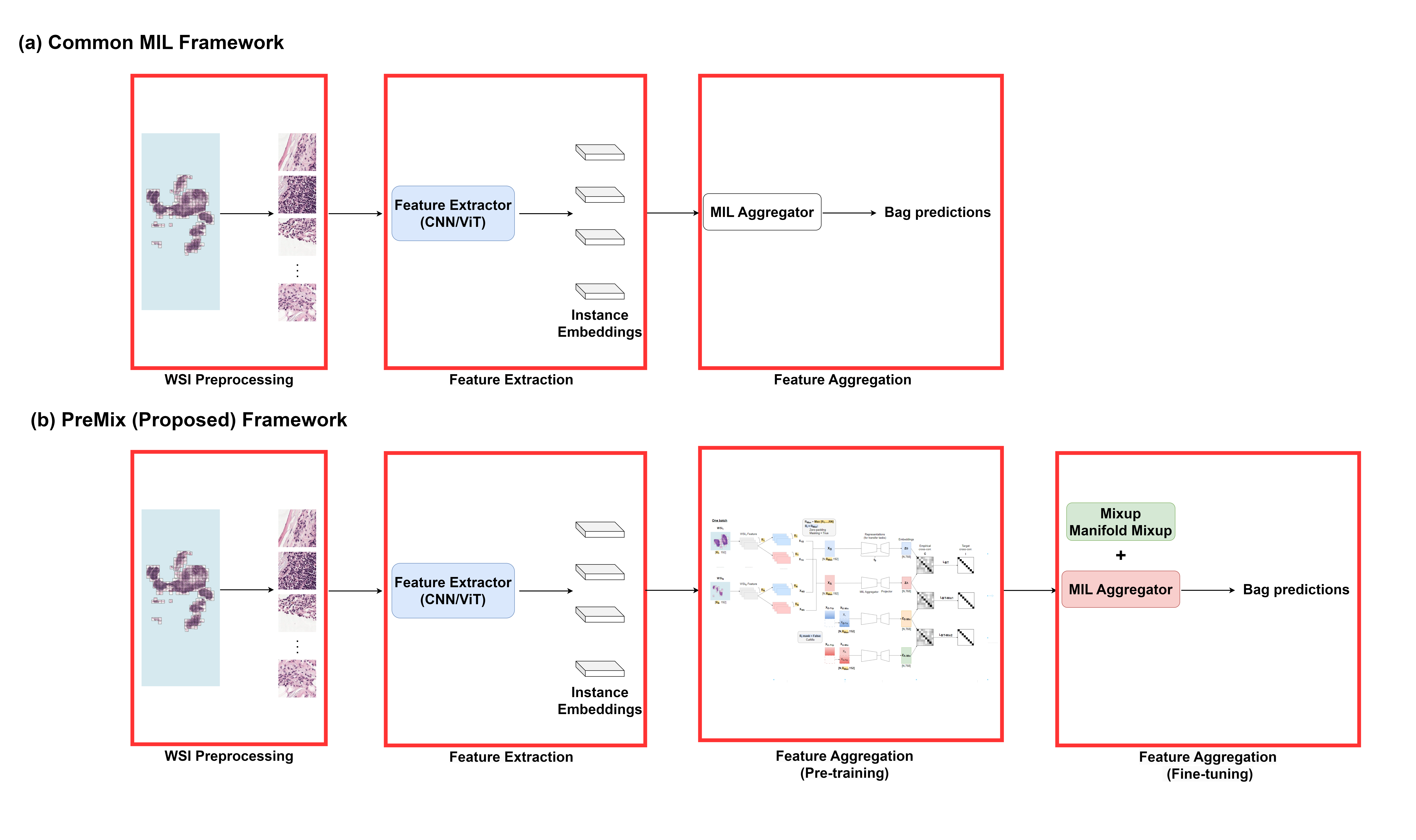}
    \caption[Common MIL vs PreMix Framework]{(a) In the common MIL framework, instance embeddings are extracted from the pre-trained feature extractor and the MIL aggregator is trained from scratch to get bag predictions. (b) In our proposed framework, PreMix, the MIL aggregator is pre-trained using unlabeled WSIs with \textit{Barlow Twins Slide Mixing} (see \hyperref[subsubsec:barlow twins slide mixing]{Section \ref{subsubsec:barlow twins slide mixing}}) and further fine-tuned with Mixup and Manifold Mixup (see \hyperref[subsubsec:mixup and manifold mixup]{Section \ref{subsubsec:mixup and manifold mixup}}). For a fair comparison, we use the original HIPT as the baseline and evaluate the integration of HIPT into PreMix.}
    \label{fig:common_mil_vs_premix_framework}
\end{figure*}

As shown in \hyperref[fig:common_mil_vs_premix_framework]{Figure \ref{fig:common_mil_vs_premix_framework}(a)}, the MIL pipeline typically begins by dividing WSIs into smaller patches, which are then processed by a pre-trained feature extractor such as a CNN \cite{ResNet} or ViT \cite{ViT}. This results in either scalar scores in instance-based MIL or feature embeddings in embedding-based MIL, with the latter demonstrating better performance in WSI classification due to its ability to capture richer and more expressive representations \citep{Embedding_vs_Instance_MIL}. These embeddings are subsequently combined using an MIL aggregator to generate slide-level predictions.

Although prior work has focused extensively on improving the quality of patch-level feature extractors \citep{CTransPath, UNI, CONCH, Rethinking_FE}, most MIL models \citep{MIL_RNN_Feature_Aggregator, ABMIL_Feature_Aggregator, DSMIL_Feature_Aggregator, CLAM_Slide_Efficient, Attention2Majority_Feature_Aggregator, DGMIL_Feature_Extractor_Aggregator, FRMIL_Feature_Aggregator, MicroMIL, shi2025positional, hezi2025cimil} still train the aggregator from random initialization. As a result, their performance tends to rely heavily on the availability of large, labeled WSI datasets, which are often scarce in real-world clinical applications due to privacy constraints \citep{Privacy}, the rarity of certain diseases \citep{Rare_Diseases1, Rare_Diseases2}, and the high cost of obtaining weak labels from domain experts \citep{Label_Efficient_1, Label_Efficient_2}. On average, an experienced pathologist needs over 20 to 30 minutes to examine and assign weak labels to a single slide \citep{qu2025weakly}, making large-scale curation highly resource intensive.

To address these limitations, a few studies have explored pre-training strategies for MIL aggregators. For example, SSMIL \citep{Contrastive_MIL_Unsupervised_Framework} applies a self-supervised contrastive learning approach using SimCLR \citep{SimCLR_SSL} at the WSI level. However, these methods have not consistently outperformed training from scratch. One possible reason is the inherent class imbalance in most WSI datasets, where negative slides (e.g., normal tissue) are much more prevalent than positive slides with tumors. This imbalance raises the risk of introducing noisy or misleading negative pairs, posing a major challenge for contrastive methods that depend on balanced positive and negative examples to learn meaningful representations.

As an alternative to contrastive learning methods that rely on negative pairs, we propose \textbf{PreMix} (\textbf{Pre}training and \textbf{Mix}ing), a self-supervised learning (SSL) framework at the WSI level designed to reduce sensitivity to class imbalance. PreMix adopts Barlow Twins \citep{Barlow_Twins_SSL}, a non-contrastive method that learns representations using \textbf{only positive pairs,} thereby avoiding issues related to negative sample selection. To further improve the effectiveness of pre-training, we introduce \textit{Barlow Twins Slide Mixing}, an intra-batch mixing strategy that interpolates features across slides within the same batch to create new positive pairs. This encourages semantic diversity and enhances representation quality, even in the absence of labeled data.

To complement the PreMix pre-training stage, we further enhance performance during fine-tuning by incorporating Mixup \citep{Mixup} and Manifold Mixup \citep{Manifold_Mixup}, which are adapted to the unique challenges of gigapixel WSIs. These mixing strategies allow for effective interpolation of both features and labels across slides of varying sizes, helping the MIL aggregator generalize better under limited supervision. By integrating both unsupervised and supervised mixing strategies, PreMix provides a unified framework that strengthens representation learning throughout the training pipeline. We validate its effectiveness by applying PreMix to HIPT \citep{HIPT_Multi_Scale}, demonstrating consistent performance gains across training datasets and annotation budgets.

The main contributions of this paper are as follows:

\begin{itemize}

\item We identify the limitations of contrastive learning (e.g., SimCLR) for MIL aggregator pre-training, especially in imbalanced WSI datasets, and instead adopt non-contrastive learning (Barlow Twins), which avoids reliance on negative pairs and yields better classification performance.

\item We propose PreMix, a novel MIL framework that introduces Barlow Twins Slide Mixing, an intra-batch feature interpolation strategy that generates diverse positive pairs and enhances representation learning during pre-training.

\item To further improve downstream performance, we integrate Mixup and Manifold Mixup during MIL aggregator fine-tuning, enabling feature and label mixing across slides of varying sizes. The Transformer encoder layer is found to yield the most significant performance gains.

\item We validate PreMix across five WSI training sizes and datasets using both random sampling (traditional fully supervised fine-tuning) and five active learning strategies: Entropy \cite{Entropy_AL}, K-means++ \cite{KMeans++_AL}, Coreset \cite{Coreset_AL}, BADGE \cite{BADGE_AL}, and CDAL \cite{CDAL_AL}, demonstrating strong robustness under limited supervision.

\end{itemize}

%% file: sections/2_Related_Work.tex
\subsection{Self-supervised learning in histopathology}

In the medical domain, where expert annotations are often scarce and expensive, SSL has emerged as a powerful alternative to fully supervised approaches \citep{SSL_Medical_Image}. In histopathology, Ciga et al. \citep{SimCLR_Feature_Extractor} showed that models pre-trained with SimCLR \citep{SimCLR_SSL} can extract strong patch-level features that generalize well across various downstream tasks. Building on this idea, Chen and Krishnan \citep{DINO_Feature_Extractor} introduced ViTs \citep{ViT} into the pipeline, replacing ResNet-50 encoders \citep{ResNet} with DINO-trained backbones \citep{DINO_SSL} to better capture complex tissue structures. Kang et al. \citep{Benchmarking_SSL_Histopathology} expanded this investigation by benchmarking several SSL methods, including Barlow Twins \citep{Barlow_Twins_SSL}, SwAV \citep{SwAV_SSL}, MoCo V2 \citep{MoCoV2_SSL}, and DINO \citep{DINO_SSL}, across diverse histopathology tasks. More recently, Wong et al. \citep{Rethinking_FE} provided a practical perspective by examining how the choice of pretraining dataset, backbone architecture, and learning objective affects performance in WSI classification.

While most SSL studies in histopathology have focused on learning \textbf{patch-level representations}, less attention has been paid to the \textbf{pretraining of MIL aggregators}. One notable exception is SSMIL \citep{Contrastive_MIL_Unsupervised_Framework}, which applied SimCLR to pre-train the MIL aggregator using WSI-level inputs. However, the performance of this contrastive approach remains inferior to models trained from scratch, suggesting that standard contrastive losses may not be well suited for the class imbalance commonly found in WSI datasets. This limitation highlights the need for alternative pretraining strategies that are more robust and compatible with the MIL setting.

\subsection{Multiple instance learning for WSI analysis}

Recent progress in MIL for WSI classification has shifted from simple aggregation techniques, such as mean or max pooling, to more sophisticated approaches that learn to combine patch-level features for slide-level predictions \citep{ReMix_Data_Augmentation_MIL}. Early efforts like MIL-RNN \citep{MIL_RNN_Feature_Aggregator} attempted to aggregate sequential patch features but were limited by the inability to capture long-range dependencies, especially in slides with a large number of patches. To address this, attention-based methods have been proposed. ABMIL \citep{ABMIL_Feature_Aggregator} assigns attention weights to individual patches, while DSMIL \citep{DSMIL_Feature_Aggregator} computes instance-level scores based on their distances to a representative “critical instance.” CLAM \citep{CLAM_Slide_Efficient} enhances the feature space through instance-level clustering, and TransMIL \citep{TransMIL_Feature_Aggregator} employs transformer-based architectures to model global relationships among patches. DTFD-MIL \citep{DFTD-MIL_Slide_Efficient} tackles label sparsity by generating pseudo-bags and estimating instance probabilities within the ABMIL framework. More recently, MHIM-MIL \citep{MHIM-MIL-Feature_Aggregator} identifies challenging instances through a momentum teacher to refine attention-based MIL training.

Although effective, these methods often operate at a single resolution and may fail to capture multi-scale spatial dependencies inherent in WSIs. To address this limitation, multi-resolution MIL strategies have been introduced. DSMIL-LC \citep{DSMIL_Feature_Aggregator} improves context-awareness by concatenating features from low and high magnifications. HIPT \citep{HIPT_Multi_Scale} advances this idea by employing a hierarchical transformer that extracts fine-grained and contextual representations across multiple resolutions. Similarly, models such as MS-DA-MIL \citep{MS-DA-MIL_Multi_Scale}, H\textsuperscript{2}-MIL \citep{H2-MIL_Multi_Scale}, and DAS-MIL \citep{DAS-MIL_Multi_Scale} leverage graph-based architectures to model both spatial and scale-aware relationships among patches. More recently, HiVE-MIL \citep{HiVE-MIL} extended this line of work by integrating vision-language models (VLMs) into a hierarchical graph structure, jointly capturing intra- and inter-scale multimodal interactions between visual and textual modalities.

Among existing methods, HIPT stands out by using a hierarchical feature extractor that integrates information across different magnifications, effectively capturing both local details and global context through \textbf{efficient feature representations}. This makes HIPT particularly appealing for pretraining, especially in resource-constrained settings where GPU capacity is limited. Motivated by this, we use the HIPT feature extractor during pretraining and adopt the full HIPT architecture as our baseline to assess how well the proposed PreMix framework performs when applied as a plug-in module.

\subsection{Data augmentation}

To overcome the limitations of small datasets and improve model generalization, a wide range of data augmentation strategies have been proposed in deep learning \citep{Data_Augmentation_Survey}. Among them, Mixup \citep{Mixup} is a well-known method that mixes both inputs and their labels to create interpolated training examples, which has inspired several domain-specific variants such as Balanced Mixup \citep{Balanced_Mixup} and Stain Mixup \citep{Stain_Mixup}. These techniques have also been extended to SSL. For instance, Ren et al. \citep{Simple_Data_Mixing_SSL} showed that using mixed natural images as supplementary positive pairs during pretraining can significantly improve the performance of contrastive frameworks like MoCo V2 \cite{MoCoV2_SSL}.

While most of these augmentation methods are developed for natural images or specific medical imaging tasks, WSIs present unique challenges due to their gigapixel scale and highly variable spatial resolution. To address this, several augmentation strategies tailored for WSIs have been introduced \citep{ReMix_Data_Augmentation_MIL,RankMix_Data_Augmentation_MIL,PseMix_Data_Augmentation_MIL}. However, these methods are typically applied only during supervised learning.

In contrast, our approach integrates data mixing at both the pretraining and fine-tuning stages. During MIL aggregator pretraining, we introduce an \textit{intra-batch mixing} strategy \textbf{at the WSI level} to generate additional positive pairs, helping the model better capture intra-cohort variability. For downstream classification, we adapt Mixup \citep{Mixup} and Manifold Mixup \citep{Manifold_Mixup} to handle variable-sized WSI features and labels. Although originally developed for fixed-size natural images, we modify these techniques to suit the scale and diversity of WSIs, resulting in improved performance and robustness under limited supervision.


%% file: sections/3_Methods.tex
\subsection{General MIL framework}
\label{subsec:General MIL framework}

As illustrated in \hyperref[fig:common_mil_vs_premix_framework]{Figure \ref{fig:common_mil_vs_premix_framework}(a)}, a common MIL framework consists of two key steps after WSI preprocessing: feature extraction and feature aggregation.

\begin{figure*}[!htb]
    \centering
    \includegraphics[width=\linewidth]{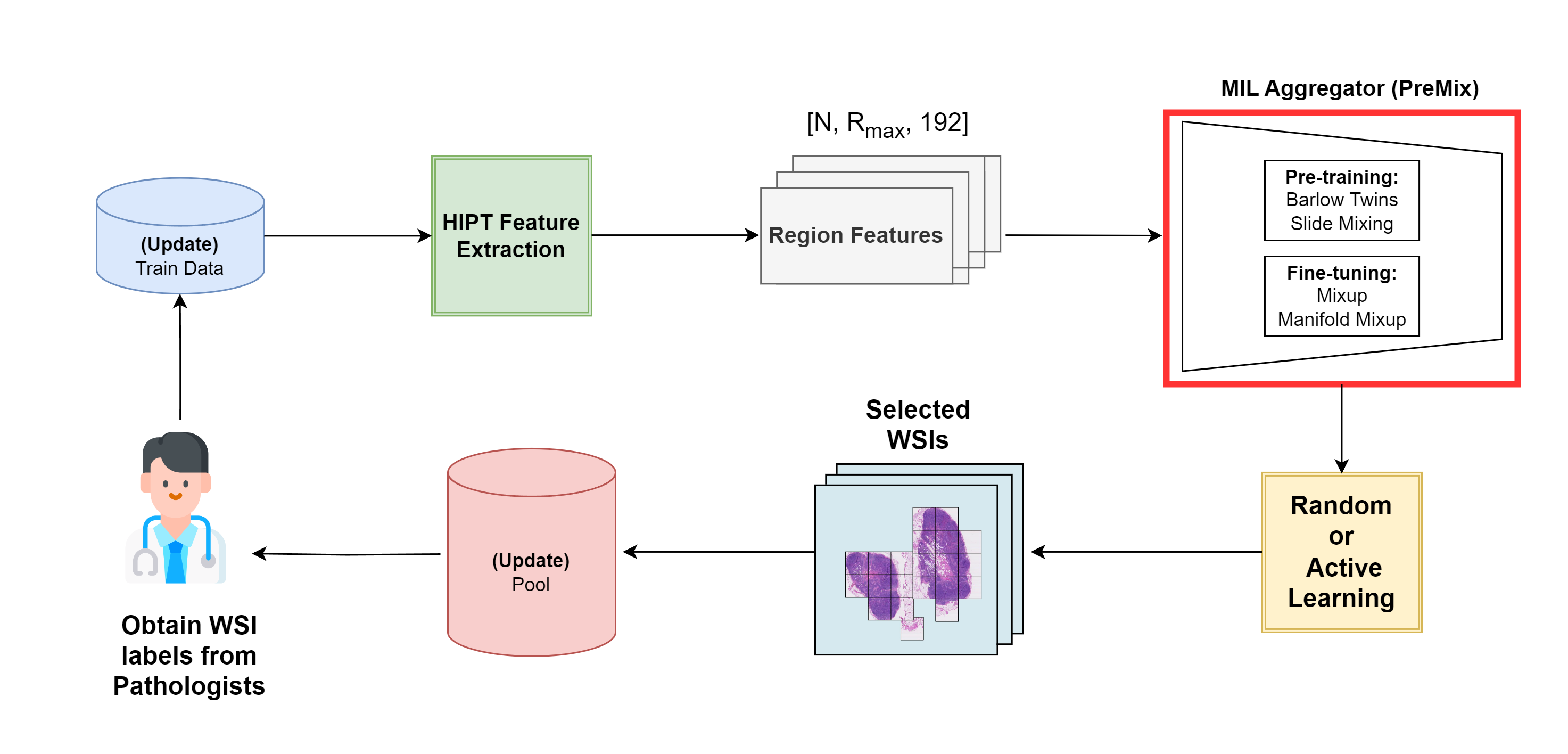}
    \caption[Overall Training Process]{The illustration of our training process utilizing the HIPT feature extraction method showcases the selection of WSI candidates for labeling using either \textbf{random sampling (traditional fully supervised fine-tuning) or active learning}. The selected WSIs are weakly-labeled by pathologists and incorporated into the next training iteration. A key distinction between the baseline HIPT and our HIPT with PreMix is that the latter incorporates MIL aggregator initialization from Barlow Twins Slide Mixing pre-training and employs Mixup and Manifold Mixup during fine-tuning, while the baseline does not.}
    \label{fig:overall_training_process}
\end{figure*}

\subsubsection{Feature extraction}
\label{subsubsec:Feature extraction}

Given a WSI denoted by the index $i$, where raw patches, typically of size $224 \times 224$ pixels, are extracted, forming the set \(X_i = \{x_{i,1}, x_{i,2}, x_{i,3}, \ldots, x_{i,n(i)}\}\). Each of these patches undergoes processing through a pre-trained patch feature extractor denoted as $f_p$, commonly implemented using architectures like ResNet \citep{ResNet} or ViT \citep{ViT}. This extraction procedure results in a set of feature representations, \(P_i = \{p_{i,1}, p_{i,2}, p_{i,3}, \ldots, p_{i,n(i)}\} \in \mathbb{R}^{n(i) \times d}\), where \(n(i)\) represents the variable number of patches extracted from the WSI $X_i$, and $d$ denotes the feature length determined by the chosen patch feature extractor.

\subsubsection{Feature aggregation}

After extraction, the individual patch features are aggregated to form a slide-level representation. This feature aggregation process is facilitated by the MIL aggregator, $f_a$. \textbf{In most cases, $f_a$ is trained from scratch (random initialization).} This process yields a comprehensive slide-level representation, expressed as $f_a(P_i)$ $\in$ $\mathbb{R}^{1 \times d}$. The primary objective of this step is to conduct the binary classification of slides. Consequently, the final representations are fed into a linear classifier to distinguish between two classes. Each slide is assigned a label $Y_i \in \{0, 1\}$, where a slide is labeled as $Y_i = 0$ if it is normal and $Y_i = 1$ if it contains tumor.

\subsection{HIPT feature extraction}
\label{subsec:HIPT feature extraction}

Contrary to most MIL methods \cite{DSMIL_Feature_Aggregator, CLAM_Slide_Efficient, FRMIL_Feature_Aggregator, DGMIL_Feature_Extractor_Aggregator}, which predominantly focus on single-level feature extraction (see \hyperref[subsubsec:Feature extraction]{Section \ref{subsubsec:Feature extraction}}), our experiment employs hierarchical visual token representations from HIPT \citep{HIPT_Multi_Scale} across various image resolutions. This approach provides a more efficient feature set and captures spatial relationships that are beneficial for both pre-training and downstream classification phases.

In this process, each WSI undergoes tiling into multiple regions, each sized $[4096, 4096]$ pixels. Subsequently, the slides are resized to the format $[R, 256, 3, 256, 256]$ with $R$ representing the number of regions in a given slide. These resized regions are then input into the pre-trained self-supervised ViT DINO 256 (1\textsuperscript{st} stage) \citep{DINO_SSL}, resulting in slide dimensions of $[R, 256, 384]$. Here, 256 signifies the sequence length of $[16, 16]$, and 384 corresponds to the output dimension of the $[CLS]$ tokens. The outputs are subsequently reshaped to $[R, 384, 16, 16]$ and fed into the pre-trained self-supervised ViT DINO 4k (2\textsuperscript{nd} stage). This final process yields a single slide representation of $[R, 192]$.

\subsection{Overall training process}

\begin{figure*}[!htb]
    \centering
    \includegraphics[width=\textwidth]{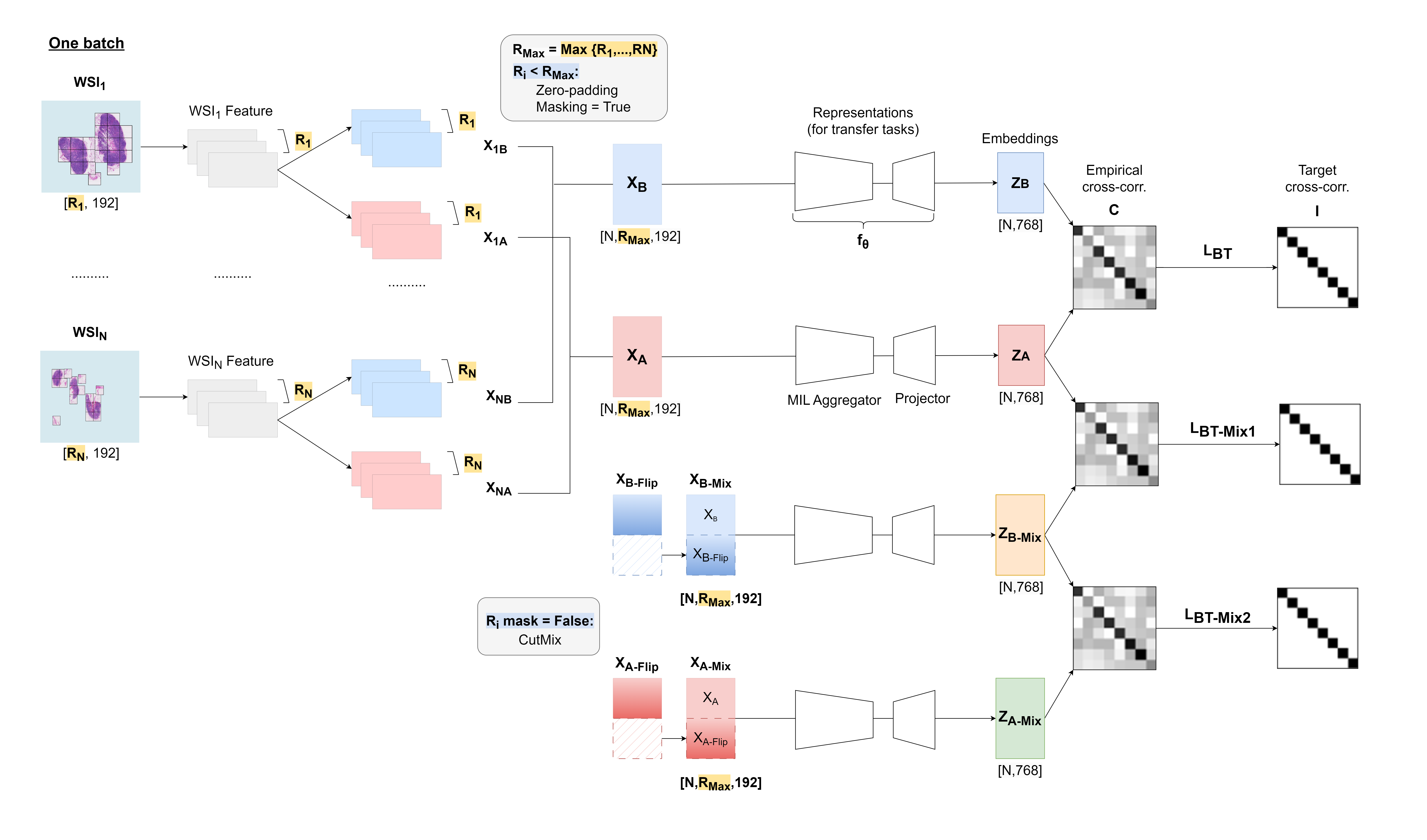}
    \caption[Barlow Twins Slide Mixing]{Barlow Twins Slide Mixing: This process relies on region features extracted via HIPT feature extractor as inputs, accommodating varying WSI input sizes through zero-padding and masking techniques and the addition of intra-batch mixing strategy during pre-training. “Flip”  means from the same batch but with the batch order reversed.}
    \label{fig:premix_barlow_twins_slide_mixing_pre-training}
\centering
\end{figure*}

In \hyperref[fig:overall_training_process]{Figure \ref{fig:overall_training_process}}, we present an overview of the comprehensive training process used in our experimental settings. Initially, we randomly select 20 WSIs, tile them into multiple regions each sized $[4096, 4096]$ pixels, and process them using the HIPT feature extraction method (see \hyperref[subsec:HIPT feature extraction]{Section \ref{subsec:HIPT feature extraction}}). The resulting dimensions are $[N, R_{max}, 192]$, where $N$ is the batch size, $R_{max}$ represents the maximum number of regions per WSI in a batch, and 192 denotes the dimensionality of the ViT DINO CLS tokens at the region level.

For the baseline, we initialize the MIL aggregator from scratch. Meanwhile, for our proposed framework, we first pre-train the MIL aggregator using the Barlow Twins Slide Mixing method (see \hyperref[subsubsec:barlow twins slide mixing]{Section \ref{subsubsec:barlow twins slide mixing}}) and then incorporate Mixup and Manifold Mixup during the fine-tuning process (see \hyperref[subsubsec:mixup and manifold mixup]{Section \ref{subsubsec:mixup and manifold mixup}}). After processing through the MIL aggregator and linear classifier, the output shape for both the baseline and proposed frameworks is $[N, num\_classes]$, representing the predicted probability for each class.

Finally, the most informative WSIs, determined by active learning acquisition functions or random sampling are selected from the unlabeled pool. These selected WSIs are then labeled by pathologists in a weakly-supervised setting, requiring only slide-level labels without fine-grained annotations. Subsequently, they are added to the training data for the next cycle. This process continues until the total labeling budget for WSIs (number of available training WSIs and their labels) is exhausted.

\subsection{PreMix}

The distinction between the original MIL framework and the PreMix lies in the pre-training MIL aggregator using Barlow Twins Slide Mixing (see \hyperref[subsubsec:barlow twins slide mixing]{Section \ref{subsubsec:barlow twins slide mixing}}) and the use of Mixup and Manifold Mixup during fine-tuning (see \hyperref[subsubsec:mixup and manifold mixup]{Section \ref{subsubsec:mixup and manifold mixup}}), whereas the original MIL framework trains the MIL aggregator directly from scratch.

\subsubsection{Pre-training: Barlow Twins Slide Mixing}
\label{subsubsec:barlow twins slide mixing}

Initializing neural networks with pre-trained models offers the advantage of leveraging knowledge from unlabeled datasets, providing a strong starting point with weights that capture general patterns and representations. This is especially effective for tasks with limited labeled data \citep{Transfer_Learning}.

The core focus of PreMix is to improve the MIL aggregator through better initialization, ultimately enhancing downstream WSI classification performance. This approach maximizes the use of unlabeled WSIs, making it especially valuable in scenarios with limited labeled data. To achieve this, we adopt the Barlow Twins method \citep{Barlow_Twins_SSL} as the base non-contrastive SSL method and introduce an \textit{intra-batch mixing} strategy to generate additional positive pairs. Since WSIs can produce varying numbers of region features, we apply zero-padding and masking techniques to ensure consistent batch dimensions during pre-training.

\textbf{Slide augmentation.} Standard augmentations from the \textit{PyTorch Torchvision} library \citep{Pytorch} are not directly applicable in our context, as our inputs are extracted WSI features rather than natural images. To address this limitation, we adapt four specific augmentations: random zeroing, Gaussian noise addition, random scaling, and random cropping, so they can be applied to WSI features during pre-training. As shown in \hyperref[fig:premix_barlow_twins_slide_mixing_pre-training]{Figure \ref{fig:premix_barlow_twins_slide_mixing_pre-training}}, for the \(N^\text{th}\) WSI in the batch, the augmented features are denoted as \(X_{N\text{A}}\) and \(X_{N\text{B}}\). After applying augmentations to all WSIs in the batch, the features \(X_{1\text{A}}\) to \(X_{N\text{A}}\) are combined into \(X_{\text{A}}\), and similarly, \(X_{1\text{B}}\) to \(X_{N\text{B}}\) are combined into \(X_{\text{B}}\).

\textbf{Intra-batch mixing.} Inspired by prior work \citep{Simple_Data_Mixing_SSL}, which demonstrated the effectiveness of image mixing in MoCo V2 for improving downstream image classification, we introduce an intra-batch mixing strategy to the Barlow Twins method (a non-contrastive approach) to enhance the generation of positive slide pairs during MIL aggregator pre-training. Specifically, slide mixing is performed within a single batch to produce additional inputs, denoted as \(X_{A-\text{Mix}}\) and \(X_{B-\text{Mix}}\), as illustrated in \hyperref[fig:premix_barlow_twins_slide_mixing_pre-training]{Figure \ref{fig:premix_barlow_twins_slide_mixing_pre-training}}.

Within the batch of WSI features, each feature indexed by \(i\) is assigned a mixing coefficient \(\lambda(i)\), sampled from a Beta distribution to determine the mixing proportion. To ensure balanced mixing, \(\lambda(i)\) is transformed to fall within the interval $[0.1, 0.9]$, resulting in a balanced mixing proportion. The modified mixing proportion, denoted as \(ratio(i)\), is then defined as:

\begin{equation}
ratio(i) = \sqrt{1 - (0.8 \times \lambda(i) + 0.1)}
\end{equation}

To identify the number of non-padding regions in the slides, we calculate:
\begin{equation}
non\_pad\_len(i) = \sum_{i} \neg mask(i)
\end{equation}

The number of regions selected for mixing, relative to the total number of non-padding regions, is determined as:
\begin{equation}
cut\_len(i) = ratio(i) \times non\_pad\_len(i)
\end{equation}

To define the boundaries of the selected regions, a random center point \(cr(i)\) is chosen, and it is computed as:
\begin{equation}
cr(i) = Rand\left( \frac{cut\_len(i)}{2}, non\_pad\_len(i) - \frac{cut\_len(i)}{2} \right)
\end{equation}

From the center point, the start and end indices of the regions are inferred as:
\begin{equation}
\begin{aligned}
    & start\_idx(i) = cr(i) - \frac{cut\_len(i)}{2}, \\
    & end\_idx(i) = cr(i) + \frac{cut\_len(i)}{2}
\end{aligned}
\end{equation}

It is crucial to ensure that the selected regions for mixing do not intersect with the masked regions. This condition is enforced by:
\begin{equation}
mask[start\_idx(i):end\_idx(i)] = \emptyset
\end{equation}

To generate the new slide-mixed feature (\(X_{A-\text{Mix}}\)), a feature \(X_A(i)\) is mixed with another feature \(X_{A-\text{Flip}}(i)\), where \(X_{A-\text{Flip}}(i)\) is obtained by reversing the batch order. This process increases the generation of positive slide pairs, providing additional inputs for pre-training compared to the original Barlow Twins method. The resulting slide-mixed feature enhances the diversity of training data and improves feature representation during pre-training.

\begin{equation}
\label{equation:mixing process}
\begin{split}
X_A[i, &\text{start\_idx}(i):\text{end\_idx}(i),:] \\
&= X_{\text{A-Flip}}[i, \text{start\_idx}(i):\text{end\_idx}(i),:]
\end{split}
\end{equation}

\textbf{Pre-training loss.} From \hyperref[equation:mixing process]{Equation~\ref{equation:mixing process}}, we derive \(X_{A-{\text{Mix}}}\) by mixing \(X_A\) and \(X_{A-\text{Flip}}\), and similarly, \(X_{B-{\text{Mix}}}\) is obtained by mixing \(X_B\) and \(X_{B-\text{Flip}}\). These inputs, \(X_A\), \(X_B\), \(X_{A-{\text{Mix}}}\), and \(X_{B-{\text{Mix}}}\), are then passed through an MIL aggregator and a projector to produce the corresponding slide embeddings \(Z_A\), \(Z_B\), \(Z_{A-{\text{Mix}}}\), and \(Z_{B-{\text{Mix}}}\).

To incorporate the mixing proportion during intra-batch mixing for loss calculation, we define \(\lambda(i)\) as:
\begin{equation}
lam(i) = 1 - \frac{cut\_len(i)}{non\_pad\_len(i)}
\end{equation}

The details of the \textsc{OriginalLoss} and \textsc{MixingLoss} functions, used to compute the Barlow Twins Slide Mixing loss, are provided in Algorithm~\ref*{algorithm:premix_barlow_twins_slide_mixing_loss_calculation}. The \textsc{OriginalLoss} function calculates the standard Barlow Twins loss, while the \textsc{MixingLoss} function computes the loss between \(Z_{B-{\text{Mix}}}\) and \(Z_A\) or \(Z_{B-{\text{Mix}}}\) and \(Z_{A-{\text{Mix}}}\), accounting for \(\lambda(i)\) as defined above. Here, “flip” refers to reversing the batch order.

\begin{algorithm}[!htb]
\small
\begin{algorithmic}[1]
\Input{Slide embeddings: $Z_A$, $Z_B$; Mixed slide embeddings: $Z_{\text{A-Mix}}$, $Z_{\text{B-Mix}}$; Mixing coefficient: $lam$}
\Output{Loss values: $source\_loss$, $mix\_source\_loss$, $mix\_loss$}
\Function{OriginalLoss}{$Z_A$, $Z_B$}
    \State $C \gets \frac{1}{\text{batch\_size}} \cdot \text{BatchNorm1d}(Z_A)^T \cdot \text{BatchNorm1d}(Z_B)$
    \State $on\_diag \gets \sum (\text{diag}(C) - 1)^2$
    \State $off\_diag \gets \sum \text{off\_diagonal}(C)^2$
    \State $loss \gets on\_diag + \lambda_{BT} \cdot off\_diag$
    \State \Return $loss$
\EndFunction

\Function{MixLoss}{$Z_A$, $Z_B$, $lam_A$, $lam_B$}
    \State $original\_loss \gets$ \Call{OriginalLoss}{$Z_A$, $Z_B$}
    \State $flip\_loss \gets$ \Call{OriginalLoss}{$Z_A$, \text{flip}($Z_B$)}
    \State $mix\_loss \gets lam_A \cdot original\_loss + lam_B \cdot flip\_loss$
    \State \Return $\text{mean}(mix\_loss)$
\EndFunction

\end{algorithmic}
\caption{Barlow Twins Slide Mixing Loss Calculation}
\label{algorithm:premix_barlow_twins_slide_mixing_loss_calculation}
\end{algorithm}

The total loss has three main components: \(source\_loss\), which is the loss between \(Z_A\) and \(Z_B\) calculated using the \textsc{OriginalLoss} function; \(mix\_source\_loss\), which is the loss between \(Z_{B-\text{Mix}}\) and \(Z_A\); and \(mix\_loss\), which represents the loss between \(Z_{B-\text{Mix}}\) and \(Z_{A-\text{Mix}}\), both computed using the \textsc{MixingLoss} function.

\begin{equation}
\begin{aligned}
\text{source\_loss} &\gets \Call{OriginalLoss}{Z_A, Z_B} \\
\text{mix\_source\_loss} &\gets \Call{MixLoss}{Z_{B-\text{Mix}}, Z_A, \lambda, 1 - \lambda} \\
\text{com} &\gets \min(\lambda, 1 - \text{flip}(\lambda)) + \min(1 - \lambda, \text{flip}(\lambda)) \\
\text{mix\_loss} &\gets \Call{MixLoss}{Z_{B-\text{Mix}}, Z_{A-\text{Mix}}, \frac{1}{1 + \text{com}}, \frac{\text{com}}{1 + \text{com}}}
\end{aligned}
\end{equation}

The total loss is defined as follows:

\begin{equation}
\label{equation:barlow twins slide mixing total loss}
total\_loss = \alpha \cdot source\_loss + \beta \cdot mix\_source\_loss + \gamma \cdot mix\_loss
\end{equation}

Where $\alpha$, $\beta$, and $\gamma$ are hyperparameters of the model that control the weight of each loss component in the total loss.

\subsubsection{Fine-tuning: Mixup and Manifold Mixup}
\label{subsubsec:mixup and manifold mixup}

\begin{figure*}[!htb]
    \centering
    \includegraphics[width=\linewidth]{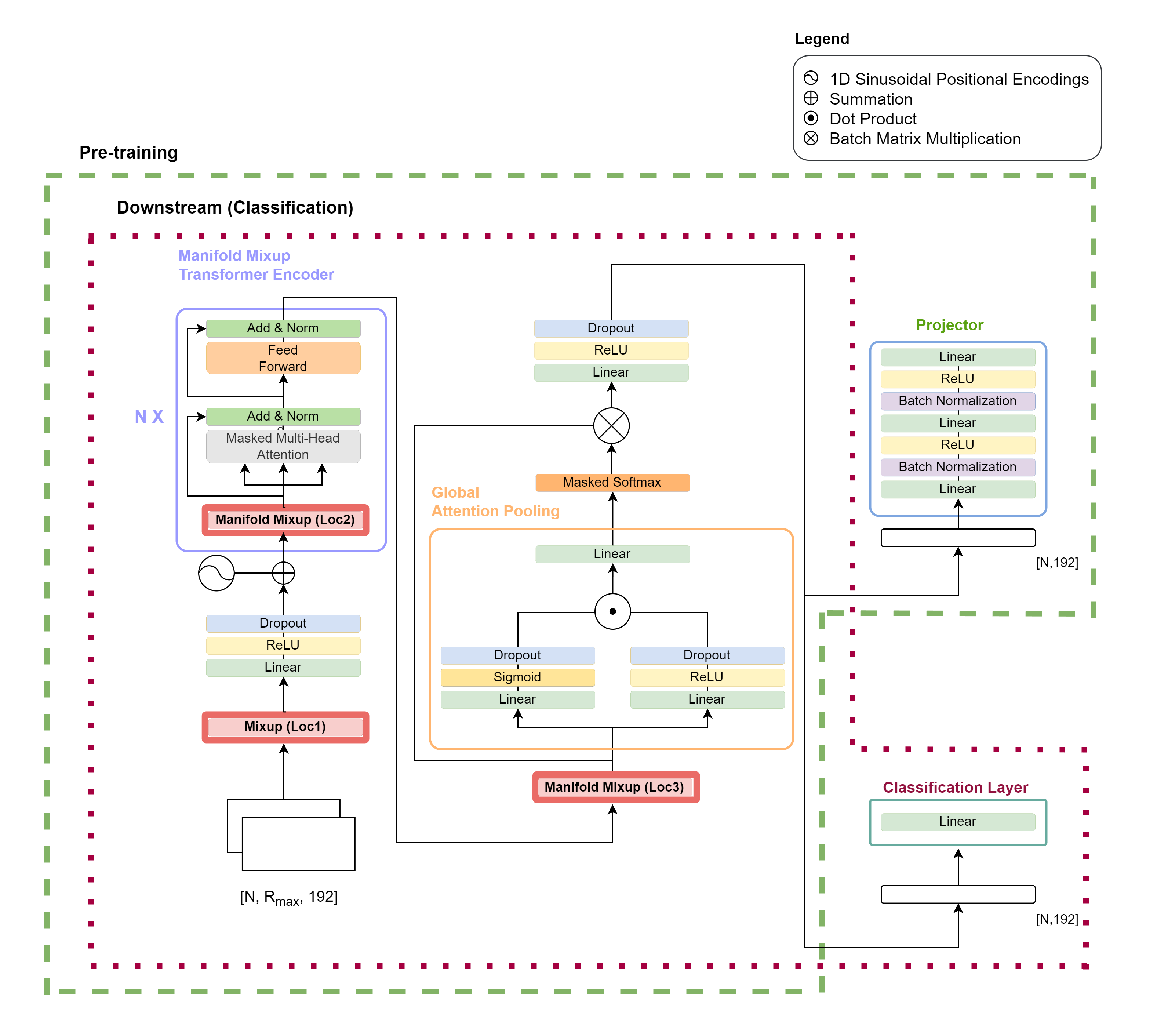}
    \caption[MIL Aggregator Architecture]{The MIL aggregator architecture supports two tasks: (1) Pre-training and (2) Downstream Classification. For pre-training, it includes components such as the Manifold Mixup Transformer Encoder, Global Attention Pooling, and Projector. For classification, it utilizes the Manifold Mixup Transformer Encoder, Global Attention Pooling, and a Classification Layer. This architecture extends the original HIPT design by incorporating positional encodings, leveraging enhanced initialization from pre-training, and employing slide feature mixing during fine-tuning (highlighted in red).}
    \label{fig:premix_mil_aggregator_architecture}
\end{figure*}

Data augmentation techniques like Mixup \citep{Mixup} and Manifold Mixup \citep{Manifold_Mixup} are widely used to enhance model performance by generating additional training samples. These methods create interpolations between data points and their labels, introducing greater variability and robustness into the training process. The primary distinction between Mixup and Manifold Mixup lies in their application: Mixup operates directly on input features, while Manifold Mixup is applied within the hidden layers of a neural network, enabling interpolations in the feature space.

In our work, we extend both Mixup and Manifold Mixup to handle the unique challenges of WSIs, where feature and label mixing involves data from varying WSI feature sizes within a single training batch. Unlike the original implementations \citep{Mixup, Manifold_Mixup}, which were designed for natural images with uniform dimensions, our approach accommodates the size variability common in WSIs. This adaptation makes the proposed method particularly effective for this domain, where feature dimensions can differ significantly between samples.

To manage the varying feature dimensions of WSIs, we standardize their shapes using zero padding and apply masking to exclude padded regions from computations. We also modify key components of the MIL aggregator, implementing masked multi-head self-attention and masked softmax operations to focus computations exclusively on non-padded regions. The detailed architecture of the MIL aggregator, including the integration points for Mixup and Manifold Mixup, is shown in \hyperref[fig:premix_mil_aggregator_architecture]{Figure \ref{fig:premix_mil_aggregator_architecture}}. 

Mixup is applied at the input stage ($Loc1$), where input features are randomly interpolated. Manifold Mixup is applied consistently within the Transformer Encoder layer ($Loc2$) and at its output ($Loc3$) to create interpolations in the feature space. These augmentations are designed to improve model robustness and generalization during fine-tuning.

%% file: sections/4_Experiments.tex
\begin{table*}[!htb]
\footnotesize
\centering
\caption{WSI dataset distribution during pre-training and downstream classification phases. An “x”  indicates that the dataset is not used in a particular process (e.g., pre-training, training, or testing), while a “-”  indicates that distributions depend on the random sampling or active learning (AL) strategy used to select a 20 WSI sample budget per iteration from the unlabeled pool.}
\label{tab:distribution of wsi datasets}
\begin{tabular}{lcc|cc|cc|cc} 
    \toprule
    \textbf{Phase} & \textbf{Dataset} & \textbf{Iteration} & \multicolumn{2}{c|}{\textbf{Pre-training}} & \multicolumn{2}{c|}{\textbf{Train}} & \multicolumn{2}{c}{\textbf{Test}} \\
    & & & \textbf{Normal} & \textbf{Tumor} & \textbf{Normal} & \textbf{Tumor} & \textbf{Normal} & \textbf{Tumor} \\ 
    \midrule
    \multirow{2}{*}{\textbf{Pre-training}} & Camelyon16 & \multirow{2}{*}{x} & 159 & 111 & \multirow{2}{*}{x} & \multirow{2}{*}{x} & \multirow{2}{*}{x} & \multirow{2}{*}{x} \\
    & CPTAC UCEC & & 185 & 390 & & & & \\
    \midrule
    \multirow{2}{*}{\textbf{Classification}} & \multirow{2}{*}{Camelyon16} & 1st (Random) & \multirow{2}{*}{x} & \multirow{2}{*}{x} & 12 & 8 & \multirow{2}{*}{80} & \multirow{2}{*}{49} \\
    & & 2nd - 5th (Rand or AL, Budget: 20) & & & - & - & & \\
    \bottomrule
\end{tabular}
\end{table*}

\subsection{Data acquisition and preparation}

This study utilizes the Camelyon16 dataset \citep{Camelyon16_Dataset}\footnote{\href{https://ftp.cngb.org/pub/gigadb/pub/10.5524/100001_101000/100439/CAMELYON16}{Camelyon16 dataset}}, which contains 399 hematoxylin and eosin (H\&E) stained whole slide images (WSIs) of breast cancer lymph node sections collected from two hospitals in the Netherlands: Radboud University Medical Center (RUMC) and University Medical Center Utrecht (UMCU). The dataset is divided into 270 training slides (159 normal and 111 tumor) and 129 test slides, serving as a widely used benchmark for WSI classification tasks.

For the pre-training phase, we supplement Camelyon16 with data from the Clinical Proteomic Tumor Analysis Consortium (CPTAC) Uterine Corpus Endometrial Carcinoma (UCEC) dataset \citep{CPTAC_Dataset}\footnote{\href{https://wiki.cancerimagingarchive.net/display/Public/CPTAC+Pathology+Slide+Downloads}{CPTAC UCEC dataset}}. From this dataset, 575 H\&E stained WSIs of endometrial tissue samples are selected, comprising 185 normal and 390 tumor slides. The UCEC dataset focuses on endometrial cancer, a significant gynecological malignancy, providing diverse tissue samples for robust representation learning. During pre-training, both Camelyon16 and UCEC datasets are utilized in an unsupervised manner, with no access to slide-level labels to ensure a purely self-supervised learning process. All test data are excluded from pre-training to prevent information leakage.

To evaluate the model's performance in data-limited scenarios, we restrict training to 100 WSIs from the Camelyon16 dataset. Initially, 20 WSIs are randomly selected for training, followed by iterative sample selection using either random sampling (traditional fully supervised fine-tuning) or active learning acquisition functions \citep{Entropy_AL, KMeans++_AL, Coreset_AL, BADGE_AL, CDAL_AL}. The complete Camelyon16 test set of 129 WSIs is retained for final evaluation. A detailed summary of the WSI dataset distribution across the pre-training and downstream classification phases is provided in \hyperref[tab:distribution of wsi datasets]{Table \ref{tab:distribution of wsi datasets}}.

\subsection{Implementation setup}

To operate under resource constraints and demonstrate the efficiency of our framework, all training experiments, including pre-training and downstream classification, are conducted on a single NVIDIA GeForce RTX 2080 Ti GPU. The models are implemented and trained using the \textit{PyTorch} library \citep{Pytorch}, a widely used open-source deep learning platform, with all experiments performed in Python. For tiling WSIs, we use the \textit{OpenSlide} library \citep{OpenSlide}.

\subsubsection{Region extraction}

WSIs are tiled into regions of size $[4096, 4096]$ pixels with no overlap between adjacent regions. A tissue threshold of 0.1 is applied to exclude non-informative regions. To standardize tissue patch extraction and address magnification differences across scanners in the Camelyon16 dataset, we employ HS2P \citep{HS2P_Preprocessing}, based on CLAM \citep{CLAM_Slide_Efficient}, with a pixel-spacing set to 0.5~μm/px (equivalent to 20$\times$ magnification).

\subsubsection{Pre-training}
\label{subsubsection:pre-training}

Pre-training is conducted using the LARS optimizer \citep{LARS_Optimizer} with a batch size of 32 over 700 epochs. The hyperparameters are adopted from the original Barlow Twins framework \citep{Barlow_Twins_SSL}, including a lambda value of $\lambda_{BT}=0.0051$, a base learning rate of 0.2, and 0.0048 for biases and batch normalization parameters. The learning rate is scaled by the batch size and normalized to 256, with a 10-epoch warm-up period followed by a cosine decay schedule that reduced the learning rate by a factor of 1000. A weight decay parameter of $10^{-6}$ is applied consistently throughout the training process.

To improve robustness during pre-training, we apply modified augmentations designed for WSI input features. These included random horizontal flipping, random zeroing, Gaussian noise addition, random scaling, and random cropping. Each augmentation is applied with a probability of 50\%, except for Gaussian noise, which is applied with a probability of 10\%. For intra-batch mixing, coefficients $\lambda$ are sampled from a Beta$(1,1)$ distribution, ensuring variation among samples within each batch. The loss function's hyperparameters $\alpha$, $\beta$, and $\gamma$ are set to 1, 0.5, and 0.5, respectively.

\begin{table*}[!htb]
\centering
\caption[Effect of Different Pre-training Initialization]{Comparison of the baseline and MIL aggregator initialization using various pre-training methods, evaluated with the F1 score.}
\label{tab:effect of different pre-training initialization}
\begin{minipage}{0.48\textwidth}
\centering
\begin{tabular}{l|lllll|l} 
    \hline
    \# Train WSIs & 20 & 40 & 60 & 80 & 100 & Mean \\ 
    \hline
    \multicolumn{7}{l}{\textbf{HIPT (Baseline)}} \\ 
    \hline
    Random & 65.1 & 65.9 & 66.7 & 65.9 & 74.4 & 67.6 \\
    Entropy & 65.1 & 68.2 & 76.7 & 72.9 & 72.9 & 71.2 \\
    BADGE & 65.1 & 64.3 & 67.4 & 70.5 & 72.1 & 67.9 \\
    Coreset & 65.1 & 64.3 & 71.3 & 79.1 & 79.8 & 71.9 \\
    K-Means++ & 65.1 & 64.3 & 72.1 & 74.4 & 71.3 & 69.5 \\
    CDAL & 65.1 & 70.5 & 70.5 & 74.4 & 76.7 & 71.5 \\
    \hline
    Mean & 65.1 & 66.3 & 70.8 & 72.9 & 74.5 & 69.9 \\
    \hline
    \hline
    \multicolumn{7}{l}{\textbf{+ SimCLR}} \\
    \hline
    Random & 62.8 & 62.8 & 64.3 & 62.8 & 68.2 & 64.2 \\
    Entropy & 62.8 & 67.4 & 67.4 & 64.3 & 74.4 & 67.3 \\
    BADGE & 62.8 & 55.8 & 67.4 & 70.5 & 69.0 & 65.1 \\
    Coreset & 62.8 & 67.4 & 66.7 & 65.9 & 65.1 & 65.6 \\
    K-Means++ & 62.8 & 55.8 & 69.0 & 65.9 & 75.2 & 65.7 \\
    CDAL & 62.8 & 63.6 & 69.0 & 70.5 & 76.7 & 68.5 \\
    \hline
    Mean & 62.8 & 62.1 & 67.3 & 66.7 & 71.4 & 66.1 \\ 
    \hline
\end{tabular}
\end{minipage}%
\hfill
\begin{minipage}{0.48\textwidth}
\centering
\begin{tabular}{l|lllll|l} 
    \hline
    \# Train WSIs  & 20 & 40 & 60 & 80 & 100 &  Mean \\ 
    \hline
    \multicolumn{7}{l}{\textbf{+ Barlow Twins}} \\
    \hline
    Random & 67.4 & 65.1 & 67.4 & 65.1 & 69.0 & 66.8 \\
    Entropy & 67.4 & 72.1 & 79.8 & 79.8 & 79.8 & \textbf{75.8} \\
    BADGE & 67.4 & 64.3 & 69.8 & 76.0 & 77.5 & 71.0 \\
    Coreset & 67.4 & 71.3 & 73.6 & 75.2 & 75.2 & 72.6 \\
    K-Means++ & 67.4 & 62.0 & 71.3 & 73.6 & 79.1 & 70.7 \\
    CDAL & 67.4 & 60.5 & 71.3 & 76.7 & 78.3 & 70.9 \\
    \hline
    Mean & 67.4 & 65.9 & 72.2 & 74.4 & 76.5 & 71.3 \\ 
    \hline
    \hline
    \multicolumn{7}{l}{\textbf{+ Barlow Twins Slide Mixing}} \\
    \hline
    Random & 69.8 & 68.2 & 73.6 & 70.5 & 72.9 & \textbf{71.0} \\
    Entropy & 69.8 & 71.3 & 74.4 & 77.5 & 78.3 & 74.3 \\
    BADGE & 69.8 & 65.9 & 73.6 & 78.3 & 78.3 & \textbf{73.2} \\
    Coreset & 69.8 & 68.2 & 77.5 & 76.0 & 79.1 & \textbf{74.1} \\
    K-Means++ & 69.8 & 72.9 & 74.4 & 74.4 & 77.5 & \textbf{73.8} \\
    CDAL & 69.8 & 67.4 & 71.3 & 76.7 & 80.6 & \textbf{73.2} \\
    \hline
    Mean & \textbf{69.8} & \textbf{69.0} & \textbf{74.2} & \textbf{75.6} & \textbf{77.8} & \textbf{73.3} \\ 
    \hline
\end{tabular}
\end{minipage}
\end{table*}

\subsubsection{Downstream classification}

Downstream classification experiments are conducted over 50 epochs with a batch size of 4. The cross-entropy loss is optimized using the Adam optimizer \citep{Adam_Optimizer}, with a learning rate of $2 \times 10^{-4}$ and a weight decay of $10^{-5}$. A StepLR scheduler is employed, with a step size of 50 epochs and a decay factor of 0.5. The training process starts with an initial subset of 20 WSIs and adds 20 more in each of the subsequent five iterations. For experiments involving Mixup and Manifold Mixup, mixing coefficients $\lambda$ were sampled from a Beta$(1,1)$ distribution and applied uniformly across all samples within each batch to ensure consistency.

\subsection{Model evaluation}

Model evaluation is conducted on the entire Camelyon16 test set, comprising 129 WSIs. The experiments utilize random sampling and five active learning acquisition functions: Entropy \citep{Entropy_AL}, K-Means++ \citep{KMeans++_AL}, Coreset \citep{Coreset_AL}, BADGE \citep{BADGE_AL}, and CDAL \citep{CDAL_AL}. Performance is assessed across five WSI training label budgets: 20, 40, 60, 80, and 100 WSIs.

In the baseline framework, the MIL aggregator is initialized and trained from scratch. To evaluate the effectiveness of the proposed framework, we compute the average F1 for each active learning acquisition function and random sampling (traditional fully supervised fine-tuning) across label budgets. This comprehensive evaluation highlights the robustness and adaptability of our framework under varying dataset compositions and labeling budgets.

\subsection{Impact of MIL aggregator pre-training methods}

Building on the findings of \citep{Contrastive_MIL_Unsupervised_Framework}, which employs the contrastive SimCLR method for MIL aggregator pre-training, we evaluate the non-contrastive Barlow Twins method alongside our proposed Barlow Twins Slide Mixing approach. The results, summarized in \hyperref[tab:effect of different pre-training initialization]{Table \ref{tab:effect of different pre-training initialization}}, reveal a clear difference in performance between these methods. Pre-training the MIL aggregator with SimCLR results in a 3.8\% reduction in mean F1 compared to the baseline (training the aggregator from scratch), averaged across random sampling (traditional fully supervised fine-tuning) and five active learning acquisition functions. 

In contrast, the vanilla Barlow Twins method yields a 1.4\% improvement over the baseline, underscoring the potential of non-contrastive approaches for WSI pretraining. Building on this, our proposed Barlow Twins Slide Mixing enhances the original method by introducing intra-batch mixing, which creates additional positive WSI pairs during training. This simple yet effective modification leads to a further 2\% gain in mean F1 score compared to the vanilla version, demonstrating the strength of our strategy for learning from unlabeled WSI data.

\subsection{Impact of Mixup and Manifold Mixup integration}

\hyperref[tab:overall results]{Table \ref{tab:overall results}} presents the average performance across random sampling and five active learning acquisition functions under five WSI label budgets. Applying Mixup and Manifold Mixup during MIL aggregator fine-tuning, following Barlow Twins Slide Mixing pretraining, achieves the best results. This setup improves mean F1 by 1.3\% over the no-mixing variant and 4.7\% over the baseline. The gains are consistent across acquisition strategies and budgets, highlighting the robustness of our framework in low-label settings. These results demonstrate the importance of effective pretraining and the benefit of incorporating feature and label mixing during fine-tuning for improved WSI classification.

\begin{table*}[!htb]
\footnotesize
\centering
\caption[Overall Results]{Overall Results: Average performance of various acquisition functions under different budget constraints, evaluated with the F1 score. “SM” means Slide Mixing.}
\label{tab:overall results}
\begin{tabular}{l|ccccc|c} 
    \hline
    \multirow{2}{*}{Methods} & \multicolumn{5}{c|}{\# WSI Training Labels} & \multirow{2}{*}{Mean} \\ 
    \cline{2-6}
    & 20 & 40 & 60 & 80 & 100 & \\
    \hline
    HIPT (Baseline) & 65.1 & 66.3 & 70.8 & 72.9 & 74.5 & 69.9 \\ 
    + Barlow Twins SM & \textbf{69.8} & 69.0 & 74.2 & 75.6 & \underline{77.8} & \underline{73.3} \\
    + Barlow Twins SM + Mixup & \underline{68.2} & 69.5 & \textbf{74.7} & 76.0 & \underline{77.8} & 73.2 \\
    + Barlow Twins SM + Mixup Manifold Mixup (PreMix) & \textbf{69.8} & \textbf{72.2} & \underline{74.5} & \textbf{78.2} & \textbf{78.4} & \textbf{74.6} \\
    \hline
    \noalign{\vskip 3pt}
    Best Improvement $\Delta$ & \textcolor{blue}{\textbf{+ 4.7}} & \textcolor{blue}{\textbf{+ 5.9}} & \textcolor{blue}{\textbf{+ 3.9}} & \textcolor{blue}{\textbf{+ 5.3}} & \textcolor{blue}{\textbf{+ 3.9}} & \textcolor{blue}{\textbf{+ 4.7}} \\
    \noalign{\vskip 3pt}
    \hline
    \end{tabular}
\end{table*}

\subsection{Ablation study}

\subsubsection{Effect of slide mixing loss hyperparameters}

We investigate the impact of adjusting the hyperparameters $(\alpha, \beta, \gamma)$ in the Barlow Twins Slide Mixing method over 300 training epochs. These hyperparameters control the relative contributions of $source\_loss$, $mix\_source\_loss$, and $mix\_loss$ as defined in \hyperref[equation:barlow twins slide mixing total loss]{Equation \ref{equation:barlow twins slide mixing total loss}}. As shown in \hyperref[tab:effect of adjusting slide mixing loss hyperparameters]{Table \ref{tab:effect of adjusting slide mixing loss hyperparameters}}, the optimal configuration is $(1, 0.5, 0.5)$, which yields the highest mean F1 score. Other settings lead to slightly lower performance, indicating that Barlow Twins Slide Mixing maintains stable effectiveness even when the loss weights are moderately varied.

\begin{table*}[!htb]
\centering
\caption[Effect of Adjusting Slide Mixing Loss Hyperparameters during pre-training]{Effect of adjusting hyperparameters $(\alpha, \beta, \gamma)$ in Barlow Twins Slide Mixing loss, corresponding to $source\_loss$, $mix\_source\_loss$, and $mix\_loss$ in \hyperref[equation:barlow twins slide mixing total loss]{Equation \ref{equation:barlow twins slide mixing total loss}}, respectively, evaluated with the F1 score.}
\label{tab:effect of adjusting slide mixing loss hyperparameters}
\begin{tabular}{p{1cm} p{1cm} p{1cm} | ccccl | l}
    \hline
    \multicolumn{3}{c|}{+ Barlow Twins Slide Mixing} & \multicolumn{5}{c|}{\# WSI Training Labels} & \multirow{2}{*}{Mean}  \\
    \cline{1-8}
    \multicolumn{1}{c}{\makebox[1cm]{$\alpha$}} & \multicolumn{1}{c}{\makebox[1cm]{$\beta$}} & \multicolumn{1}{c|}{\makebox[1cm]{$\gamma$}} & 20 & 40 & 60 & 80 & 100 & \\
    \hline
    \multicolumn{1}{c}{0} & \multicolumn{1}{c}{0.5} & \multicolumn{1}{c|}{0.5} & 65.9 & 66.5 & 73.3 & 76.4 & 77.3 & 71.9 \\
    \multicolumn{1}{c}{0} & \multicolumn{1}{c}{1} & \multicolumn{1}{c|}{1} & 67.4 & 69.9 & 73.5 & 74.9 & 76.6 & 72.5 \\
    \multicolumn{1}{c}{1} & \multicolumn{1}{c}{0.5} & \multicolumn{1}{c|}{0.5} & 66.7 & 70.2 & 72.9 & 76.7 & 77.6 & \textbf{72.8} \\
    \multicolumn{1}{c}{1} & \multicolumn{1}{c}{1} & \multicolumn{1}{c|}{1} & 70.5 & 68.0 & 74.0 & 73.8 & 76.4 & 72.5 \\
    \hline
\end{tabular}
\end{table*}

\subsubsection{Effect of Mixup and Manifold Mixup locations}

We evaluate the impact of applying Mixup and Manifold Mixup at different stages of MIL aggregator fine-tuning. Specifically, we assess performance when mixing is applied at $Loc1$, $Loc2$, and $Loc3$, as shown in \hyperref[fig:premix_mil_aggregator_architecture]{Figure \ref{fig:premix_mil_aggregator_architecture}}. As reported in \hyperref[tab:effect of mixing locations in downstream classification]{Table \ref{tab:effect of mixing locations in downstream classification}}, applying mixing at all three locations yields the highest mean F1 score of 74.6\%, outperforming all other configurations. In contrast, using only $Loc1$ leads to a slight performance drop, suggesting that mixing at a single stage is insufficient. Notably, applying Manifold Mixup at $Loc2$, within the Transformer Encoder, results in a 2.5\% gain in mean F1, indicating that this layer plays a key role in improving classification accuracy.

\begin{table*}[!htb]
\centering
\caption[Effect of Mixing Locations in Downstream Classification] {Effect of Mixup and Manifold Mixup at different combination locations when fine-tuning the MIL aggregator, evaluated with the F1 score. Refer to \hyperref[fig:premix_mil_aggregator_architecture]{Figure \ref{fig:premix_mil_aggregator_architecture}} for details on $Loc1$, $Loc2$, and $Loc3$.}
\label{tab:effect of mixing locations in downstream classification}
\begin{tabular}{p{1cm} p{1cm} p{1cm} | ccccl | l}
    \hline
    \multicolumn{3}{c|}{+ Barlow Twins Slide Mixing} & \multicolumn{5}{c|}{\# WSI Training Labels} & \multirow{2}{*}{Mean}  \\
    \cline{1-8}
    \multicolumn{1}{c}{\makebox[1cm]{$Loc1$}} & \multicolumn{1}{c}{\makebox[1cm]{$Loc2$}} & \multicolumn{1}{c|}{\makebox[1cm]{$Loc3$}} & 20 & 40 & 60 & 80 & 100 & \\
    \hline
    \multicolumn{1}{c}{\cmark} & \multicolumn{1}{c}{\xmark} & \multicolumn{1}{c|}{\xmark} & 68.2 & 69.5 & 74.7 & 76.0 & 77.8 & 73.2 \\
    \multicolumn{1}{c}{\cmark} & \multicolumn{1}{c}{\xmark} & \multicolumn{1}{c|}{\cmark} & 65.9 & 70.7 & 73.4 & 74.3 & 76.1 & 72.1 \\
    \multicolumn{1}{c}{\cmark} & \multicolumn{1}{c}{\cmark} & \multicolumn{1}{c|}{\cmark} & 69.8 & 72.2 & 74.5 & 78.2 & 78.4 & \textbf{74.6} \\
    \hline
\end{tabular}
\end{table*}

%% file: sections/5_Discussion.tex
Most existing methods train the MIL aggregator from scratch, often overlooking the valuable information available in unlabeled WSIs. Our findings show that fine-tuning the MIL aggregator using contrastive learning methods such as SimCLR results in poorer performance compared to supervised training, which is consistent with previous studies \citep{Contrastive_MIL_Unsupervised_Framework}. The main drawback of contrastive learning lies in its reliance on both positive and negative pairs. In the context of WSI classification, where there is often a large imbalance between normal and tumor slides, this reliance increases the risk of including incorrect or noisy negative samples. To overcome this issue, we adopt non-contrastive methods such as Barlow Twins, which require only positive pairs. Unlike contrastive approaches, our experiments demonstrate that non-contrastive pre-training significantly improves WSI classification performance.

We further extend the Barlow Twins method through our proposed Barlow Twins Slide Mixing, which introduces intra-batch mixing to create additional positive slide pairs during pre-training. By combining features from different slides within the same batch, this approach enhances semantic diversity and allows the MIL aggregator to learn more robust and generalizable representations.

To improve performance during fine-tuning, we apply Mixup and Manifold Mixup in a way that accounts for the unique properties of WSIs, including their large size and varying patch counts. These techniques interpolate both features and labels across slides, generating synthetic training examples that improve generalization. They also help address class imbalance by producing mixed slides that better represent underrepresented classes.

Our ablation study shows that the performance of our method remains stable across a range of hyperparameter values for slide mixing loss. This reduces the need for extensive hyperparameter tuning and makes the method more practical. In addition, we find that the location where mixing is applied within the MIL aggregator matters. In particular, applying Manifold Mixup within the Transformer Encoder layer yields the largest performance gain, likely due to the richer and more abstract representations learned at this stage. This location allows the model to benefit from interpolated features while preserving the contextual structure of the original slides.

We validate our framework under different labeling conditions and training set sizes. Under traditional fully supervised fine-tuning with random sampling, our method consistently outperforms the baseline. We also evaluate it in an active learning setting using five acquisition strategies: Entropy, K-means++, Coreset, BADGE, and CDAL. These strategies select the most informative slides for training, resulting in different subsets of the data. Across all conditions, our method remains robust and effective, highlighting its adaptability and reliability in limited supervision settings.

\textbf{Limitation and Future Work.} While our method improves performance, the pre-training dataset is limited in organ diversity and slide count. In future work, we plan to expand the dataset to include a broader range of organs and cancer types to enhance generalizability.

%% file: sections/6_Conclusions.tex
MIL aggregator pre-training is a crucial yet underexplored area in weakly supervised WSI classification. Most existing approaches train MIL aggregators from scratch, overlooking the potential of unlabeled WSIs. This paper introduces PreMix, a framework designed to leverage extensive, underutilized unlabeled WSI datasets by employing Barlow Twins Slide Mixing for MIL aggregator pre-training. By incorporating Mixup and Manifold Mixup during fine-tuning, PreMix demonstrates robust performance across diverse WSI training datasets and labels, including traditional fully supervised fine-tuning settings and five active learning techniques. The results highlight PreMix's potential to enhance WSI classification with limited labeled data and its applicability to real-world histopathology practices.